\documentclass{article} 
\usepackage{iclr2016_conference,times}
\usepackage{hyperref}
\usepackage{url}
\usepackage[pdftex]{graphicx}

\title{Convolutional networks and learning invariant
to homogeneous multiplicative scalings}

\author{Mark Tygert, Arthur Szlam, Soumith Chintala, Marc'Aurelio Ranzato,\\
{\bf Yuandong Tian \& Wojciech Zaremba}\\
Facebook Artificial Intelligence Research\\
\texttt{\{tygert,aszlam,soumith,ranzato,yuandong,wojciech\}@fb.com}}

\begin{document}

\maketitle

\begin{abstract}
The conventional classification schemes --- notably multinomial logistic
regression --- used in conjunction with convolutional networks (convnets)
are classical in statistics, designed without consideration
for the usual coupling with convnets, stochastic gradient descent,
and backpropagation. In the specific application to supervised learning
for convnets, a simple scale-invariant classification stage turns out to be
more robust than multinomial logistic regression, appears to result
in slightly lower errors on several standard test sets,
has similar computational costs,
and features precise control over the actual rate of learning.
``Scale-invariant'' means that multiplying the input values
by any nonzero real number leaves the output unchanged.
\end{abstract}

\section{Introduction}

Classification of a vector of real numbers (called ``feature activations'')
into one of several discrete categories is well established and well studied,
with generally satisfactory solutions such as the ubiquitous
multinomial logistic regression reviewed, for example,
by~\citet{hastie-tibshirani-friedman}.
However, the canonical classification may not couple well
with generation of the feature activations via convolutional networks
(convnets) trained using stochastic gradient descent,
as discussed, for example, by~\citet{lecun-bottou-bengio-haffner}.
Fitting (also known as learning or training) the combination
of the convnet and the classification stage by minimizing
the cost/loss/objective function associated with the classification
suggests designing a stage specifically for use
in such joint fitting/learning/training.
In particular, the convnets presented in the present paper are ``equivariant''
with respect to scalar multiplication --- multiplying the input values
by any real number multiplies the output by the same factor;
the present paper leverages this equivariance
via a ``scale-invariant'' classification stage
--- a stage for which multiplying the input values by any nonzero real number
leaves the output unchanged.
The scale-invariant classification stage turns out to be more robust to
outliers (including obviously mislabeled data), fits/learns/trains precisely
at the rate that the user specifies, and apparently results
in slightly lower errors on several standard test sets
when used in conjunction with some typical convnets
for generating the feature activations.
The computational costs are comparable to those
of multinomial logistic regression.
Similar classification has been introduced earlier in other contexts
by~\citet{hill-doucet}, \citet{lange-wu}, \citet{wu-lange},
\citet{saberian-vasconcelos}, \citet{mroueh-poggio-rosasco-slotine},
\citet{wu-wu}, and others.
Complementary normalization includes the work of~\citet{carandini-heeger},
\citet{ioffe-szegedy}, and the associated references.
The key to effective learning is rescaling, as described
in Section~\ref{new} below (see especially the last paragraph there).
This rescaled learning, while necessary for training convolutional networks,
is unnecessary in the aforementioned earlier works.

The remainder of the present paper has the following structure:
Section~\ref{notation} sets the notation.
Section~\ref{new} introduces the scale-invariant classification stage.
Section~\ref{robust} analyzes its robustness.
Section~\ref{experiments} illustrates the performance of the classification
on several standard data sets.
Section~\ref{conclusion} concludes the paper.

\section{Notational conventions}
\label{notation}

All numbers used in the classification stage will be real valued
(though the numbers used for generating the inputs to the stage
may in general be complex valued).
We follow the recommendations of~\citet{magnus-neudecker}:
all vectors are column vectors (aside from gradients
of a scalar with respect to a column vector, which are row vectors),
and we use $\|v\|$ to denote the Euclidean norm of a vector $v$;
that is, $\|v\|$ is the square root of the sum of the squares of the entries
of $v$.
We use $\|A\|$ to denote the spectral norm of a matrix $A$; that is,
$\|A\|$ is the greatest singular value of $A$, which is also
the maximum of $\|Av\|$ over every vector $v$ such that $\|v\| = 1$.
The terminology ``Frobenius norm'' of $A$ refers to the square root
of the sum of the squares of the entries of $A$.
The spectral norm of a vector viewed as a matrix having only one column
or one row is the same as the Euclidean norm of the vector;
the Euclidean norm of a matrix viewed as a vector
is the same as the Frobenius norm of the matrix.

\section{A scale-invariant classification stage}
\label{new}

We study a linear classification stage that assigns one of $k$ classes
to each real-valued vector $x$ of feature activations
(together with a measure of confidence in its classification),
with the assignment being independent of the Euclidean norm of $x$;
the Euclidean norm of $x$ is its ``scale.''
We associate to the $k$ classes target vectors $t_1$,~$t_2$, \dots, $t_k$
that are the vertices of either a standard simplex or a regular simplex
embedded in a Euclidean space of dimension $m \ge k$ ---
the dimension of the embedding space being strictly greater than the minimum
($k-1$) required to contain the simplex will give extra space
to help facilitate learning;
\citet{hill-doucet}, \citet{lange-wu}, \citet{wu-lange},
\citet{saberian-vasconcelos}, \citet{mroueh-poggio-rosasco-slotine},
and~\citet{wu-wu} (amongst others) discuss these simplices
and their applications to classification.
For the standard simplex, the targets are just the standard basis vectors,
each of which consists of zeros for all but one entry.
For both the regular and standard simplices,
\begin{equation}
\label{unitt}
\|t_1\| = \|t_2\| = \dots = \|t_k\| = 1.
\end{equation}
Given an input vector $x$ of feature activations,
we identify the target vector $t_j$ that is nearest in the Euclidean distance
to
\begin{equation}
\label{z}
z = \frac{y}{\|y\|},
\end{equation}
where
\begin{equation}
\label{y}
y = Ax
\end{equation}
for an $m \times n$ matrix $A$ determined via learning as discussed shortly.
The index $j$ such that $\|z-t_j\|$ is minimal is the index of the class
to which we assign $x$.
The classification is known as ``linear'' or ``multi-linear'' due to~(\ref{y}).
The index $j$ to which we assign $x$ is clearly independent
of the Euclidean norm of $x$ due to~(\ref{z}),
and the assignment is ``scale-invariant''
even if we rescale $A$ by a nonzero scalar multiple.

To determine $A$, we first initialize all its entries to random numbers,
then divide each entry by the Frobenius norm of $A$
and multiply by the square root of the number of rows in $A$.
We then conduct iterations of stochastic gradient descent
as advocated by~\citet{lecun-bottou-bengio-haffner},
updating $A$ to $\tilde{A}$ on each iteration via
\begin{equation}
\label{update}
\tilde{A} = A - h \left(\frac{\partial c}{\partial A}\right)^\top,
\end{equation}
where $h$ is a positive real number (known as the ``learning rate''
or ``step length'') and $c$ is the cost to be minimized
that is associated with a vector chosen at random from among the input vectors
and its associated vector $x$ of feature activations,
\begin{equation}
\label{c}
c = \|z-t_j\|^2,
\end{equation}
where $t_j$ is the target for the correct class associated with $x$,
and $z$ is the vector-valued function of $x$ specified in~(\ref{z})
and~(\ref{y}).

As elaborated by~\citet{lecun-bottou-bengio-haffner},
usually we combine stochastic gradient descent with backpropagation
to update the entries of $x$ associated with the chosen input,
which requires propagating the gradient $\partial c/\partial x$
back into the network generating the feature activations that are the entries
of $x$ for the chosen input sample. We use the same learning rate
from the classification stage throughout the network
generating the feature activations. Fortunately, a straightforward calculation
shows that the Euclidean norm of the gradient $\partial c/\partial x$
is bounded independent of the scaling of $A$:
\begin{equation}
\label{Jacobound}
\left\| \frac{\partial c}{\partial x} \right\| \le \frac{4 \|A\|}{\|Ax\|};
\end{equation}
please note that scaling the matrix $A$ by any nonzero scalar multiple
has no effect on the right-hand side of~(\ref{Jacobound})
--- the gradient propagating in backpropagation is independent of the size
of $A$.

Critically, after every update as in~(\ref{update}), we rescale the matrix $A$:
we divide every entry by the Frobenius norm of $A$
and multiply by the square root of the number of rows in $A$.
We use the rescaled matrix for subsequent iterations
of stochastic gradient descent.
Rescaling $A$ yields precisely the same vector $z$ in~(\ref{z})
and cost $c$ in~(\ref{c}); together with the scale-invariance
of the right-hand side of~(\ref{Jacobound}), rescaling ensures that
the stochastic gradient iterations are effective and numerically stable
for any learning rate $h$.

\section{Robustness}
\label{robust}

Combining~(\ref{unitt}) and the fact that the Euclidean norm of $z$
from~(\ref{z}) is 1 yields that the cost $c$ from~(\ref{c}) satisfies
\begin{equation}
\label{c2}
0 \le c \le 4.
\end{equation}
As reviewed by~\citet{hastie-tibshirani-friedman},
the cost associated with classification via multinomial logistic regression is
\begin{equation}
\label{r2}
r = \ln\left(\sum_{\ell=1}^m \exp(y^{(\ell)})\right)-y^{(j)},
\end{equation}
where $j$ is the index among $1$,~$2$, \dots, $k$ of the correct class,
and $y^{(1)}$,~$y^{(2)}$, \dots, $y^{(m)}$ are the entries of the vector $y$
from~(\ref{y}), with $m = k$ for multinomial logistic regression.
Whereas the cost $c$ is bounded as in~(\ref{c2}),
the cost $r$ from~(\ref{r2}) is bounded only for positive values of $y^{(j)}$,
growing linearly for negative $y^{(j)}$.
Thus, $c$ is more robust than $r$ to outliers;
logistic regression is less robust to outliers
(including obviously mislabeled inputs).

\section{Numerical experiments}
\label{experiments}

\begin{figure}
\begin{flushright}
(a)\parbox{.93\linewidth}{\includegraphics[width=.92\linewidth]{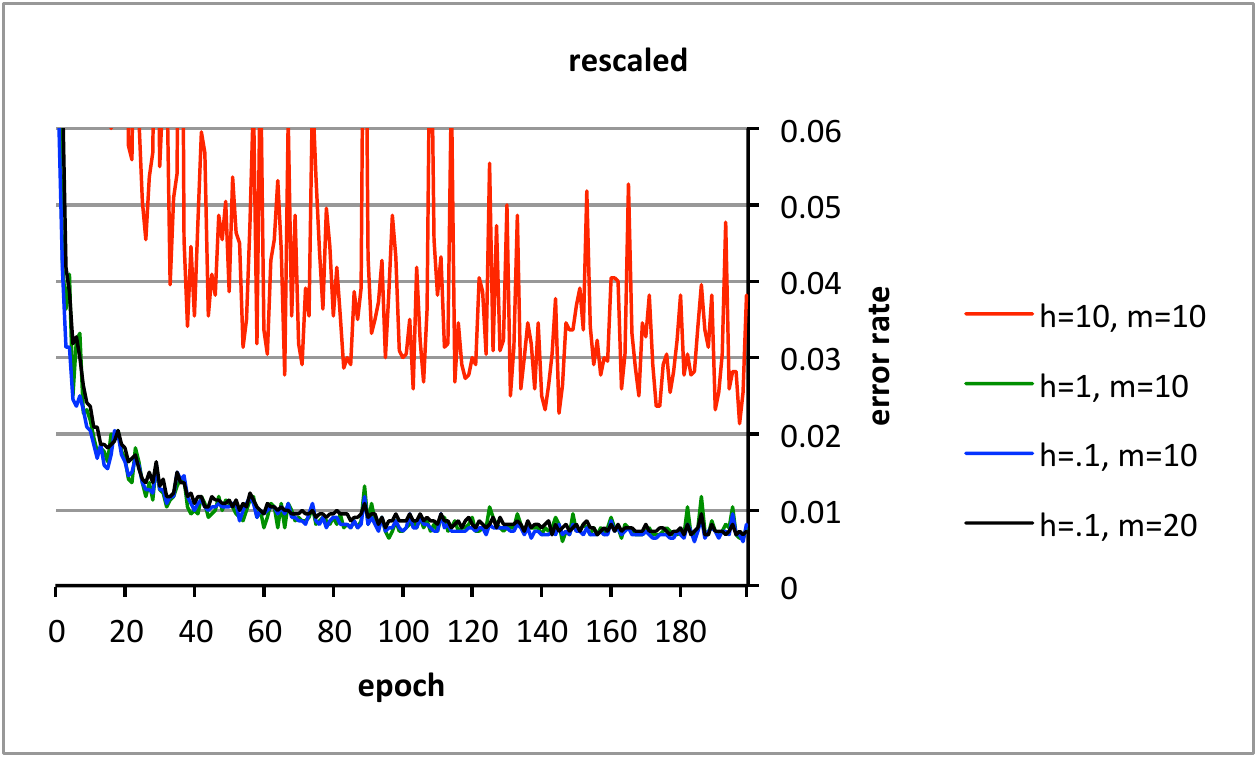}}
(b)\parbox{.93\linewidth}{\includegraphics[width=.92\linewidth]{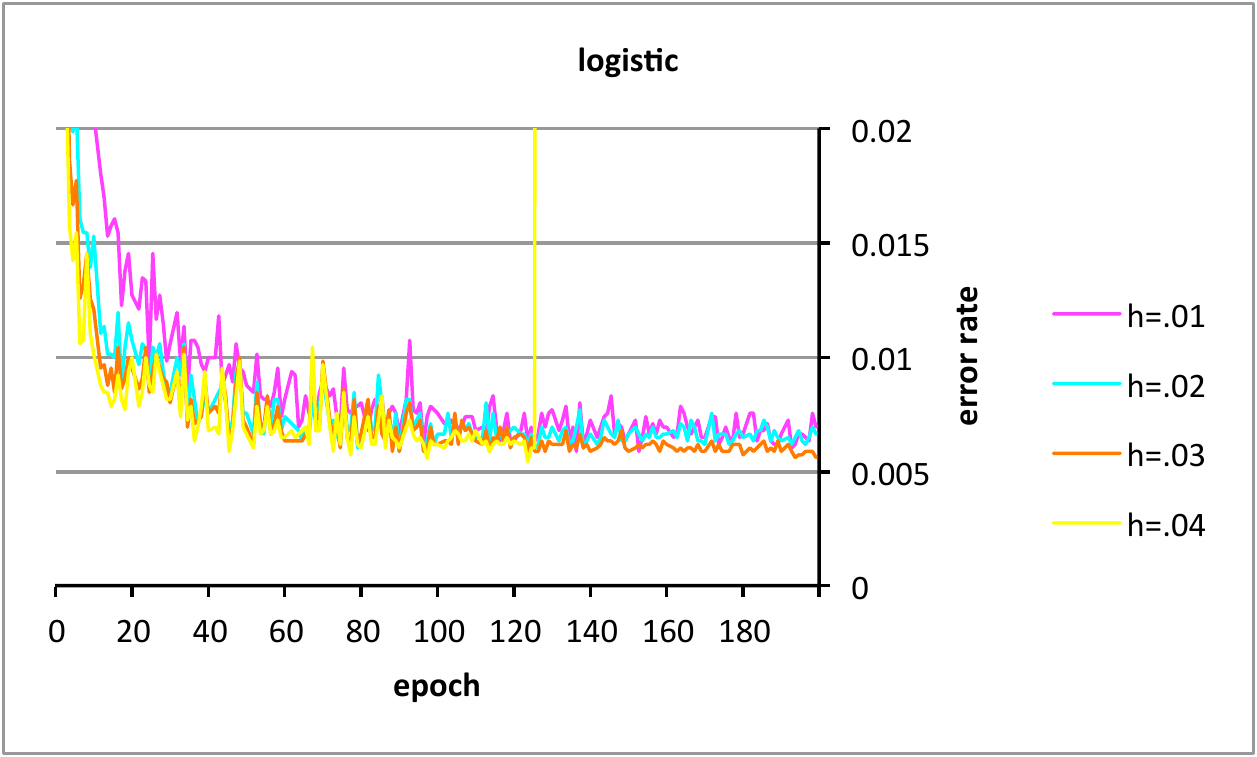}}
(c)\parbox{.93\linewidth}{\includegraphics[width=.92\linewidth]{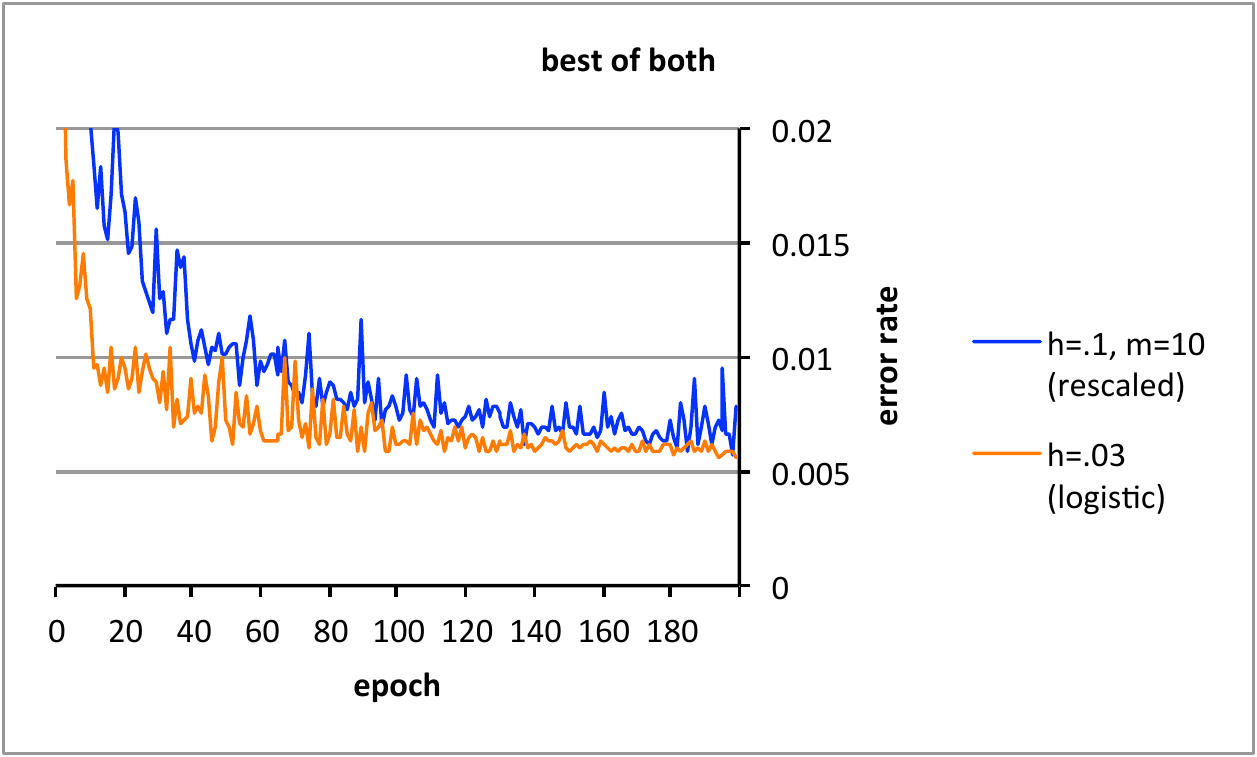}}
\end{flushright}
\caption{MNIST --- larger architecture}
\label{mfig}
\end{figure}

The present section provides a brief empirical evaluation
of rescaling in comparison with the usual multinomial logistic regression,
performing the learning for both via stochastic gradient descent
(the learning is end-to-end, training the entire network --- including
both the convolutional network and the classification stage --- jointly,
with the same learning rate everywhere).
The experiments (and corresponding figures) consider various choices
for the learning rate $h$ and for the dimension $m$ of the space
containing the simplex targets.
We renormalize the parameters in the classification stage after every minibatch
of 100 samples when rescaling (not with the multinomial logistic regression),
as detailed in the last paragraph of Section~\ref{new}
and the penultimate paragraph of the present section.
The rescaled approach appears to perform somewhat better
than multinomial logistic regression in all but Figure~\ref{mfig}
for the experiments detailed in the present section.
The remainder of the present section provides details.

Following~\citet{lecun-bottou-bengio-haffner},
the architectures for generating the feature activations
are convolutional networks (convnets) consisting of series of stages,
with each stage feeding its output into the next
(except for the last, which feeds into the classification stage).
Each stage convolves each image from its input
against several learned convolutional kernels, summing together
the convolved images from all the inputs into several output images,
then takes the absolute value of each pixel of each resulting image,
and finally averages over each patch in a partition of each image
into a grid of $2 \times 2$ patches.
All convolutions are complex valued and produce pixels
only where the original images cover all necessary inputs (that is,
a convolution reduces each dimension of the image by one less than the size
of the convolutional kernel).
We subtract the mean of the pixel values from each input image
before processing with the convnets, and we append
an additional feature activation to those obtained from the convnets,
namely the standard deviation of the set of values of the pixels in the image.
For each data set, we use two network architectures,
where the second is a somewhat smaller variant of the first.
We consider three data sets whose training properties are reasonably
straightforward to investigate, with each set consisting of $k=10$ classes
of images; the first two are the usual CIFAR-10 and MNIST
of~\citet{krizhevsky} and~\citet{lecun-bottou-bengio-haffner}.
The third is a subset of the 2012 ImageNet data set
of~\citet{imagenet2012}, retaining 10 classes of images,
representing each class by 100 samples
in a training set and 50 per class in a testing set.
CIFAR-10 contains 50,000 images in its training set
and 10,000 images in its testing set.
MNIST contains 60,000 images in its training set
and 10,000 images in its testing set.
The images in the MNIST set are grayscale.
The images in both the CIFAR-10 and ImageNet sets are full color,
with three color channels. We neither augmented the input data
nor regularized the cost/loss functions.
We used the Torch7 platform --- http://torch.ch --- for all computations.

Tables~\ref{standard}--\ref{subsetp} display the specific configurations
we used.
``Stage'' specifies the positions of the indicated layers in the convnet.
``Input images'' specifies the number of images input to the given stage
for each sample from the data.
``Output images'' specifies the number of images output from the given stage.
Each input image is convolved against a separate, learned convolutional kernel
for each output image (with the results of all these convolutions
summed together for each output image).
``Kernel size'' specifies the size of the square grid of pixels
used in the convolutions.
``Input image size'' specifies the size of the square grid of pixels
constituting each input image.
``Output image size'' specifies the size of the square grid of pixels
constituting each output image.
Tables~\ref{standard} and~\ref{standardp} display the two configurations
used for processing both CIFAR-10 and MNIST.
Tables~\ref{subset} and~\ref{subsetp} display the two configurations
used for processing the subset of ImageNet described above.

\begin{figure}
\begin{flushright}
(a)\parbox{.93\linewidth}{\includegraphics[width=.92\linewidth]{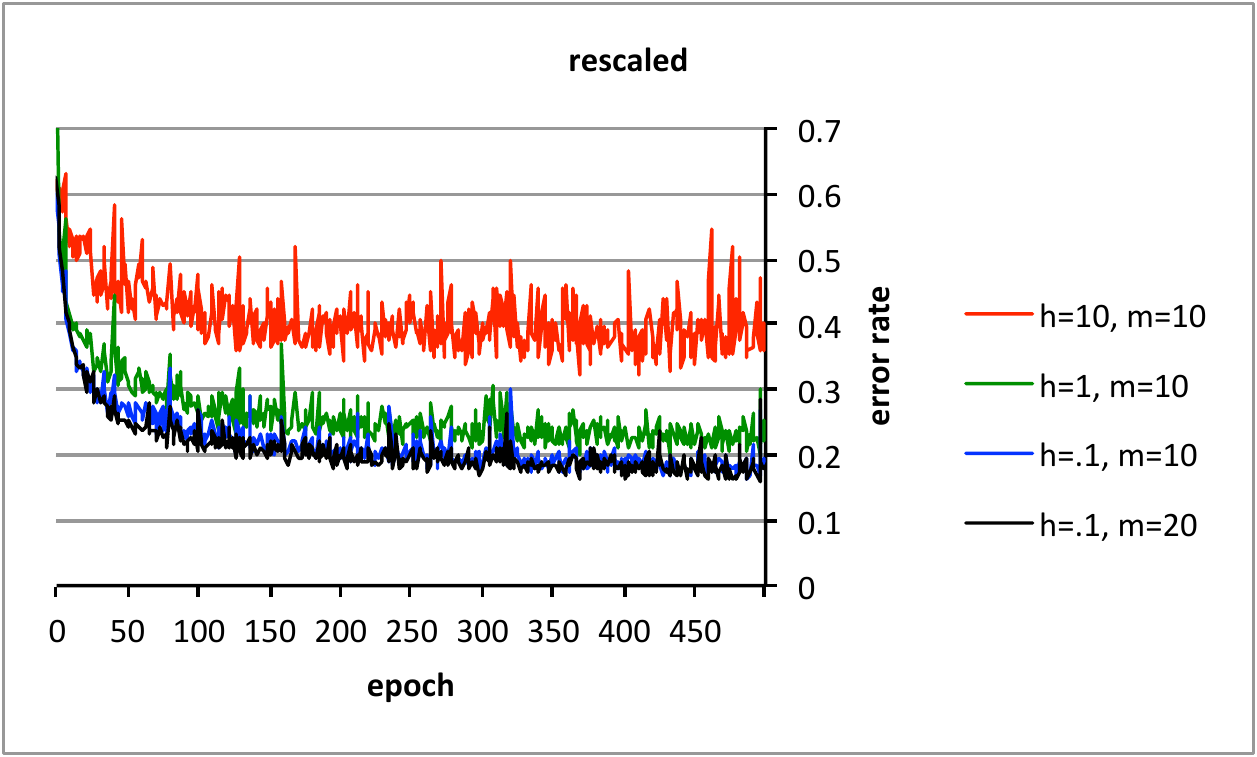}}
(b)\parbox{.93\linewidth}{\includegraphics[width=.92\linewidth]{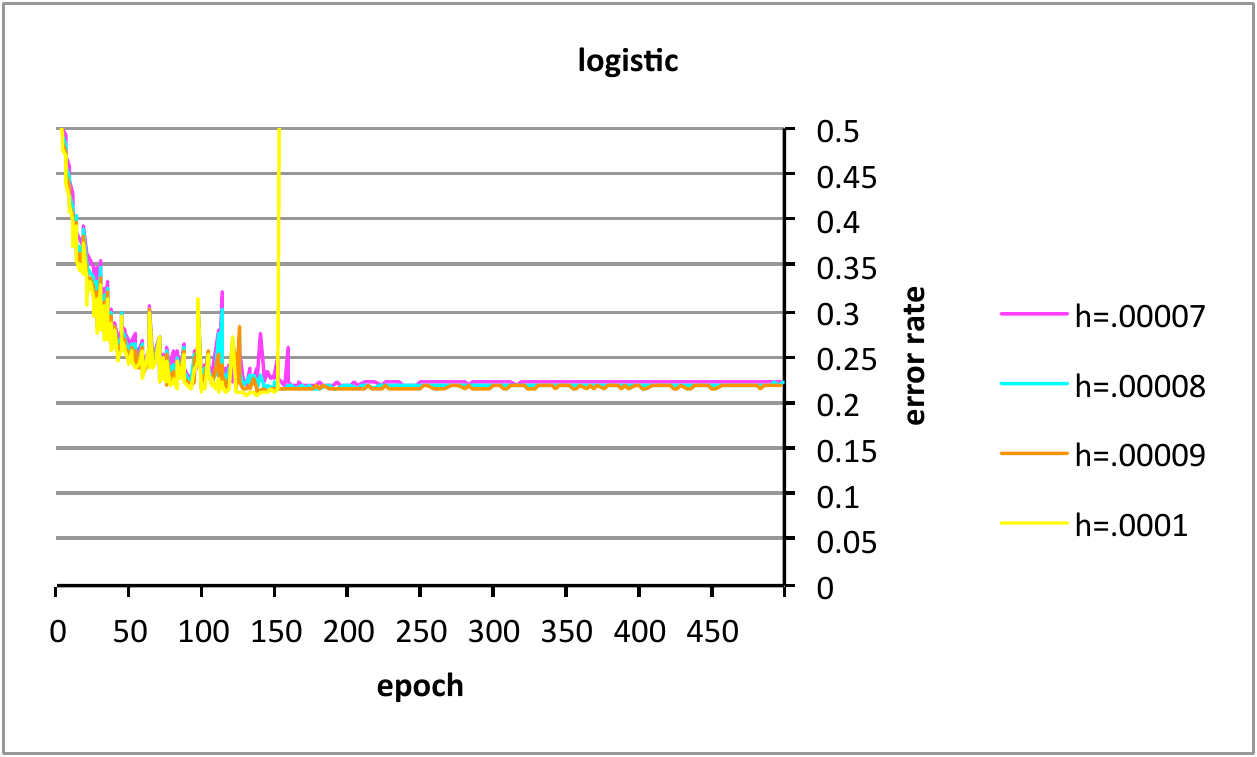}}
(c)\parbox{.93\linewidth}{\includegraphics[width=.92\linewidth]{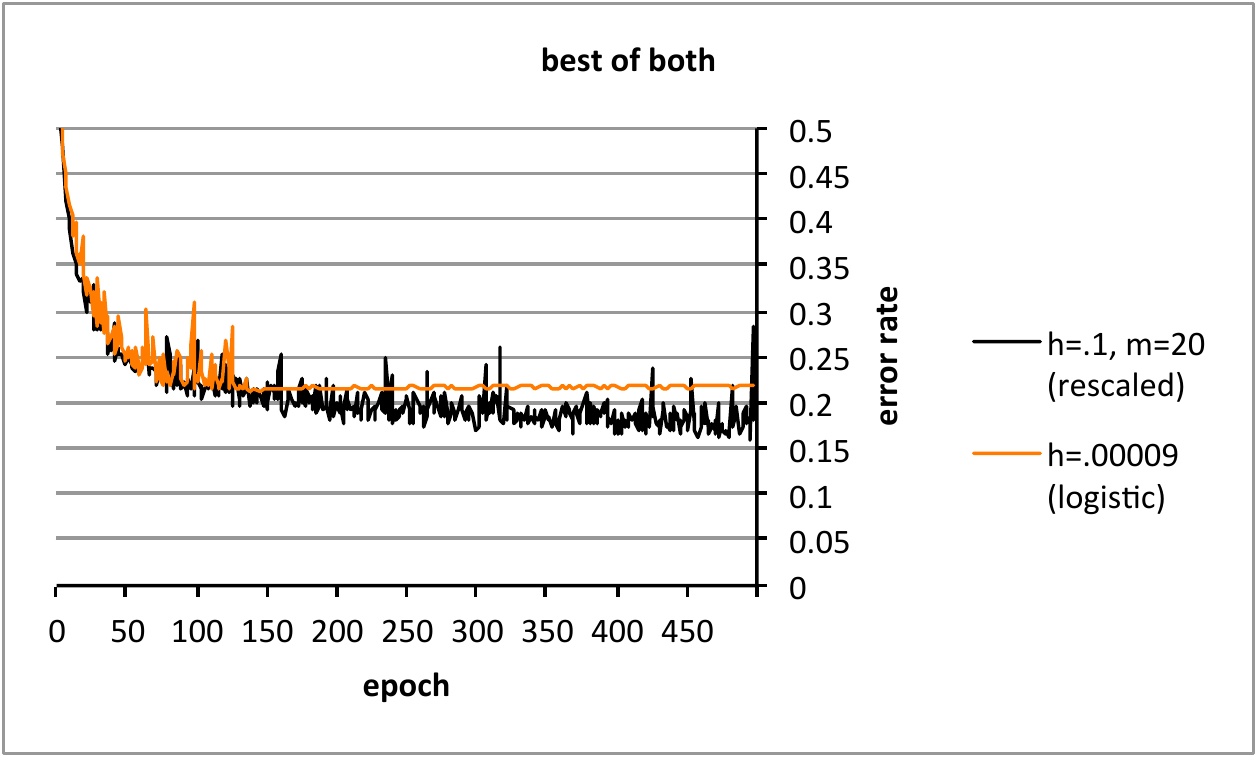}}
\end{flushright}
\caption{CIFAR-10 --- larger architecture}
\label{cfig}
\end{figure}

\begin{figure}
\begin{flushright}
(a)\parbox{.93\linewidth}{\includegraphics[width=.92\linewidth]{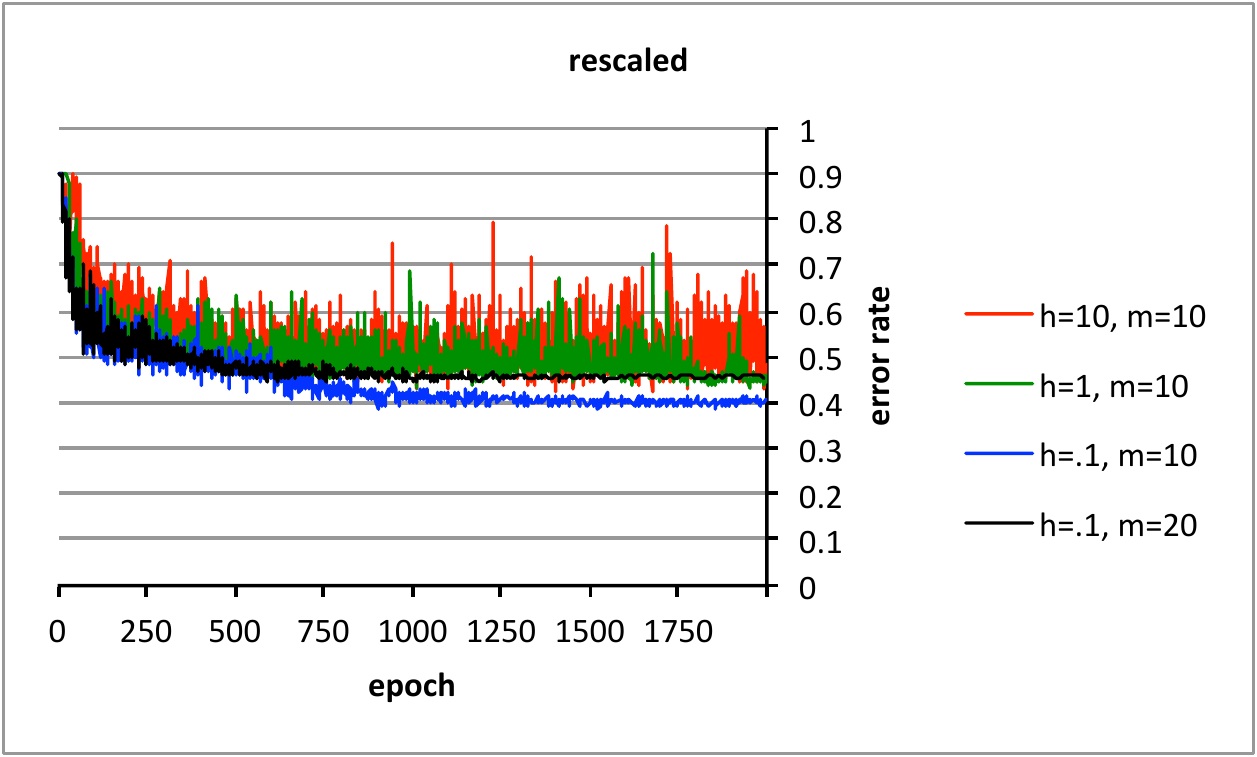}}
(b)\parbox{.93\linewidth}{\includegraphics[width=.92\linewidth]{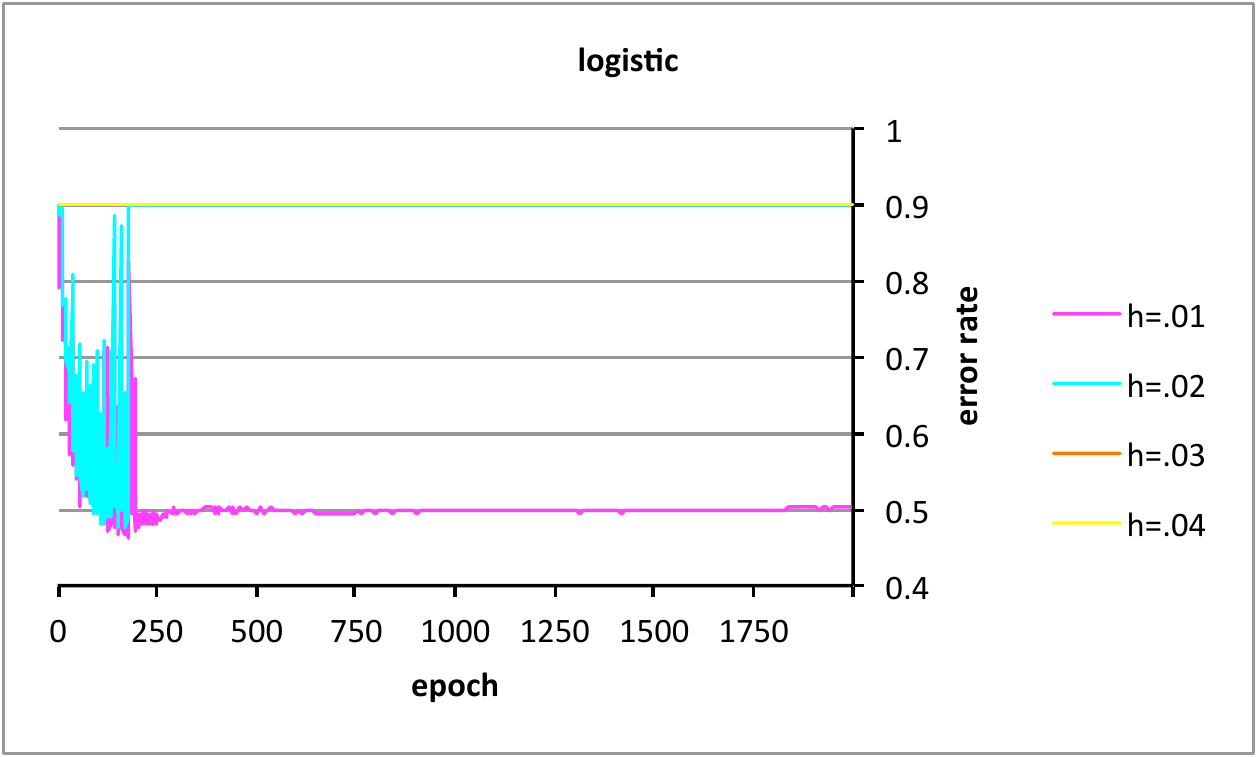}}
(c)\parbox{.93\linewidth}{\includegraphics[width=.92\linewidth]{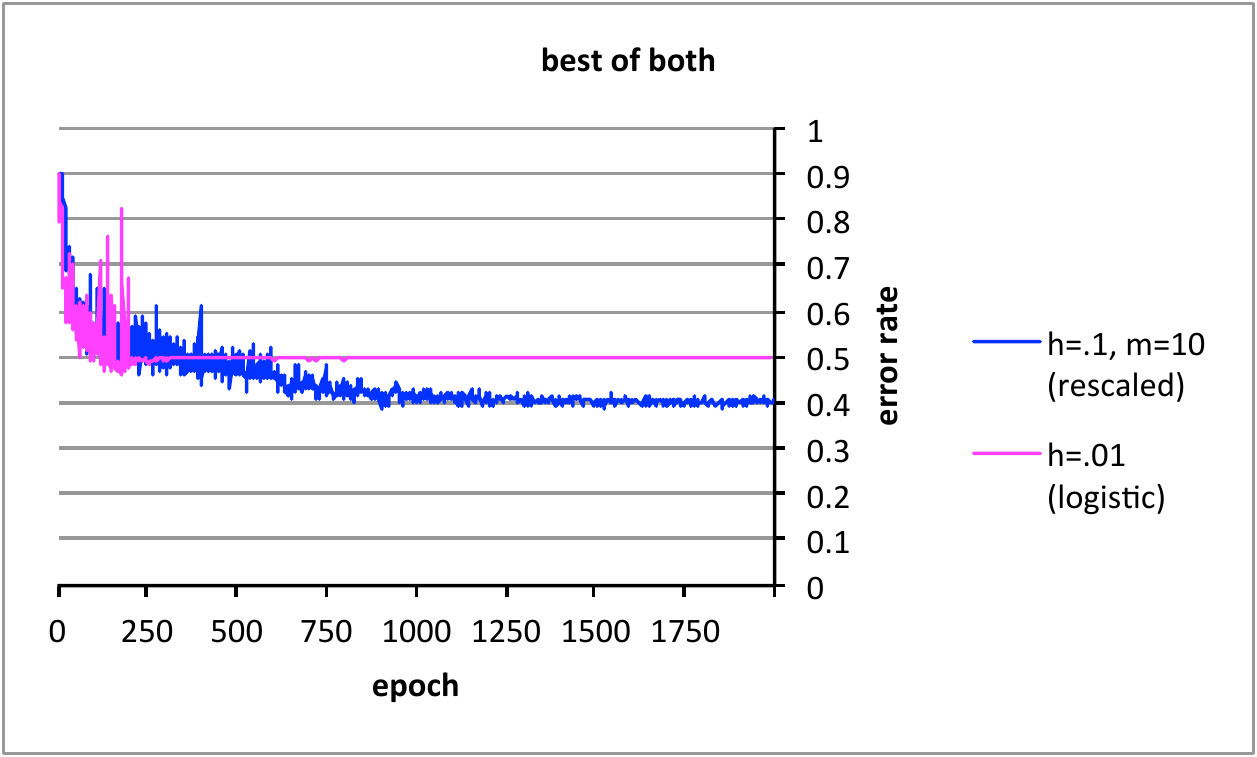}}
\end{flushright}
\caption{ImageNet subset --- larger architecture}
\label{ifig}
\end{figure}

\begin{figure}
\begin{flushright}
(a)\parbox{.93\linewidth}{\includegraphics[width=.92\linewidth]{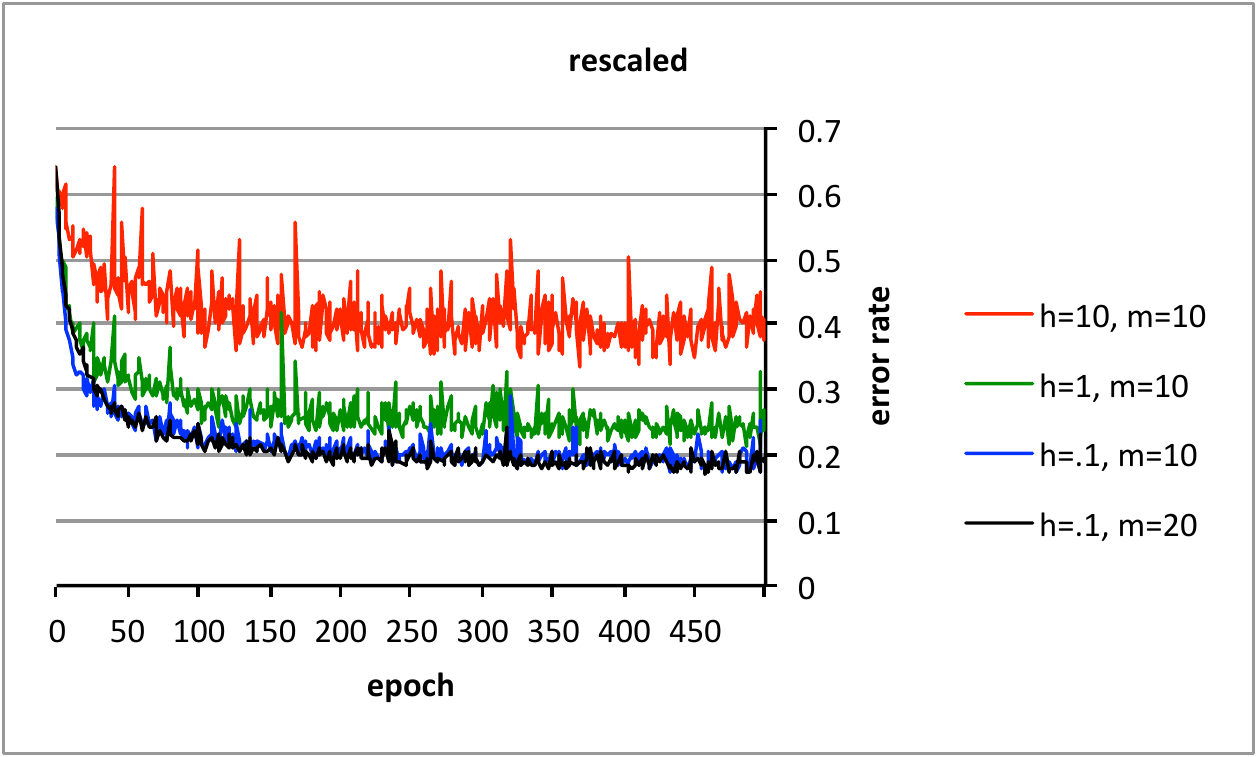}}
(b)\parbox{.93\linewidth}{\includegraphics[width=.92\linewidth]{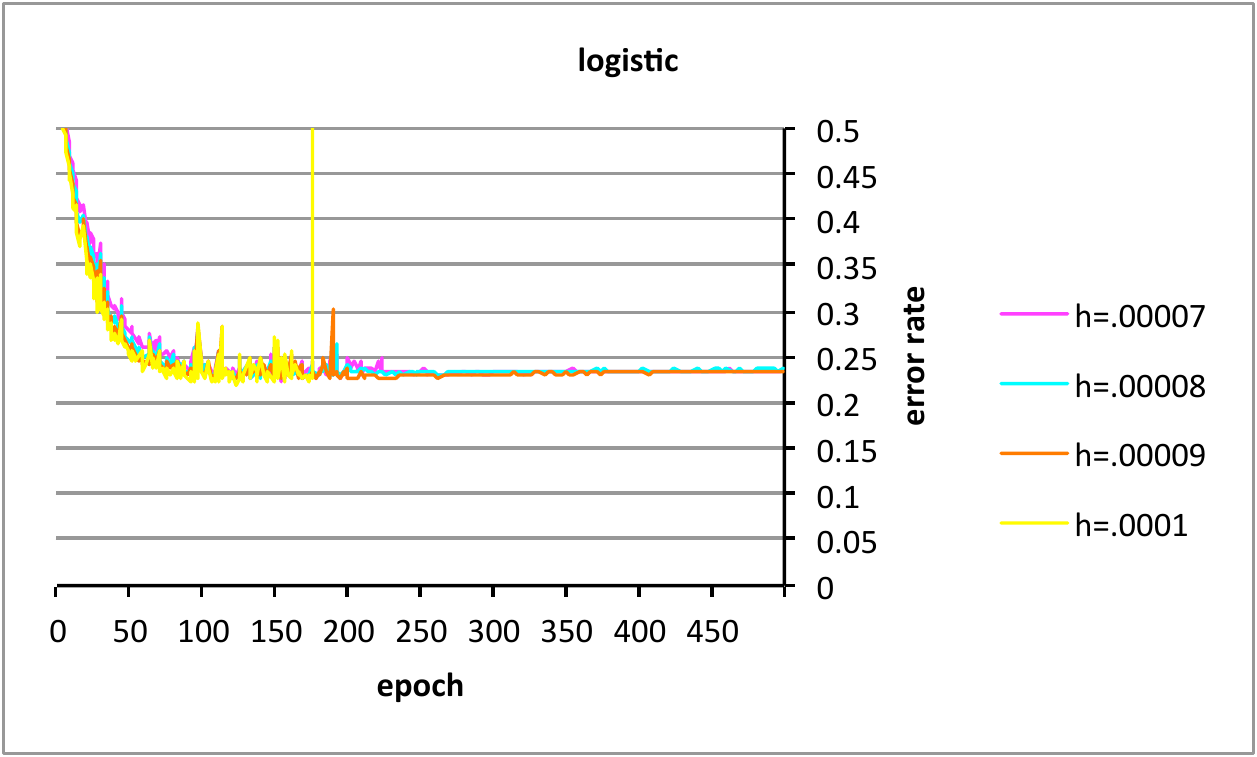}}
(c)\parbox{.93\linewidth}{\includegraphics[width=.92\linewidth]{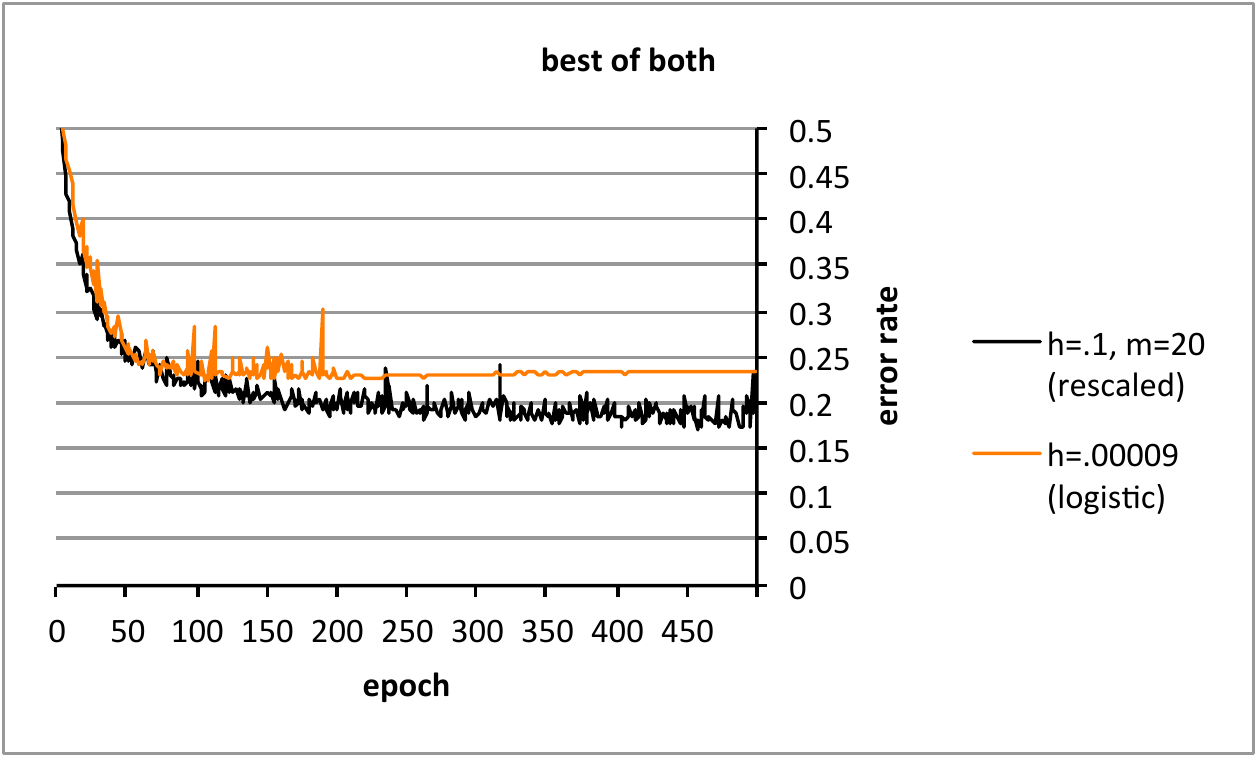}}
\end{flushright}
\caption{CIFAR-10 --- smaller architecture}
\label{cfigp}
\end{figure}

\begin{figure}
\begin{flushright}
(a)\parbox{.93\linewidth}{\includegraphics[width=.92\linewidth]{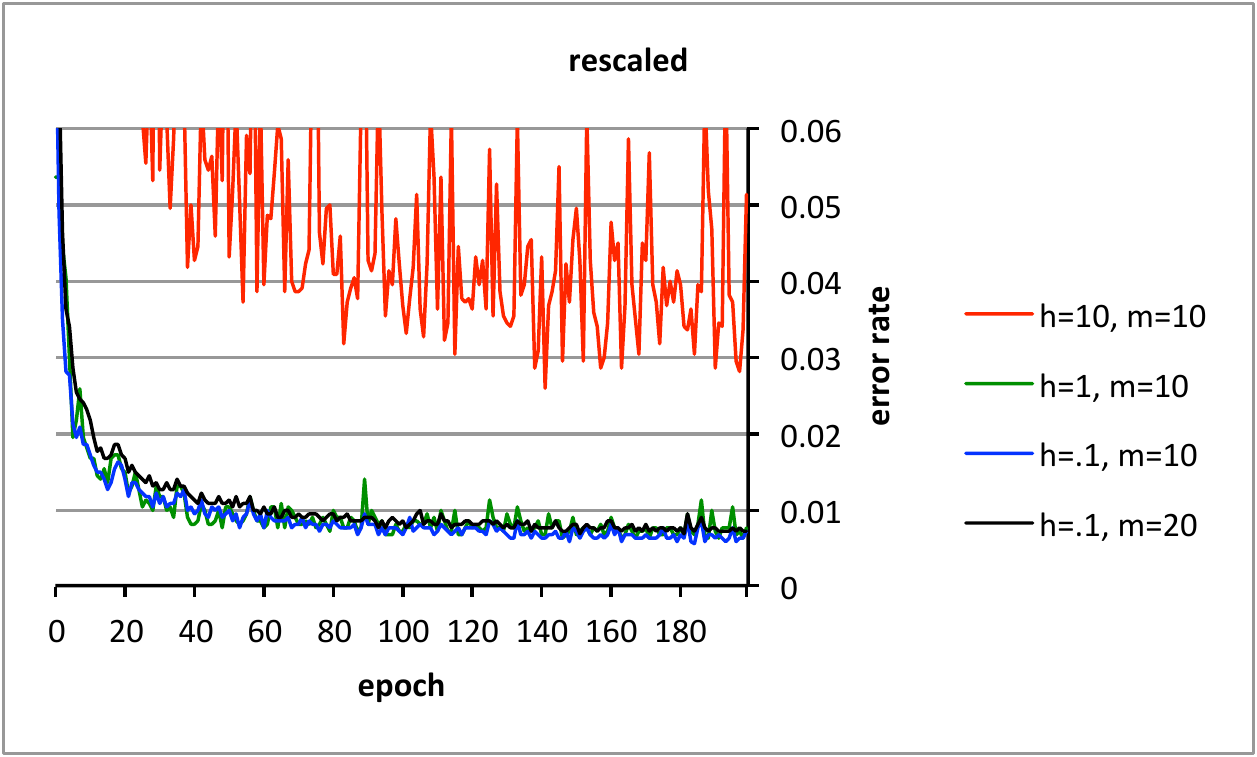}}
(b)\parbox{.93\linewidth}{\includegraphics[width=.92\linewidth]{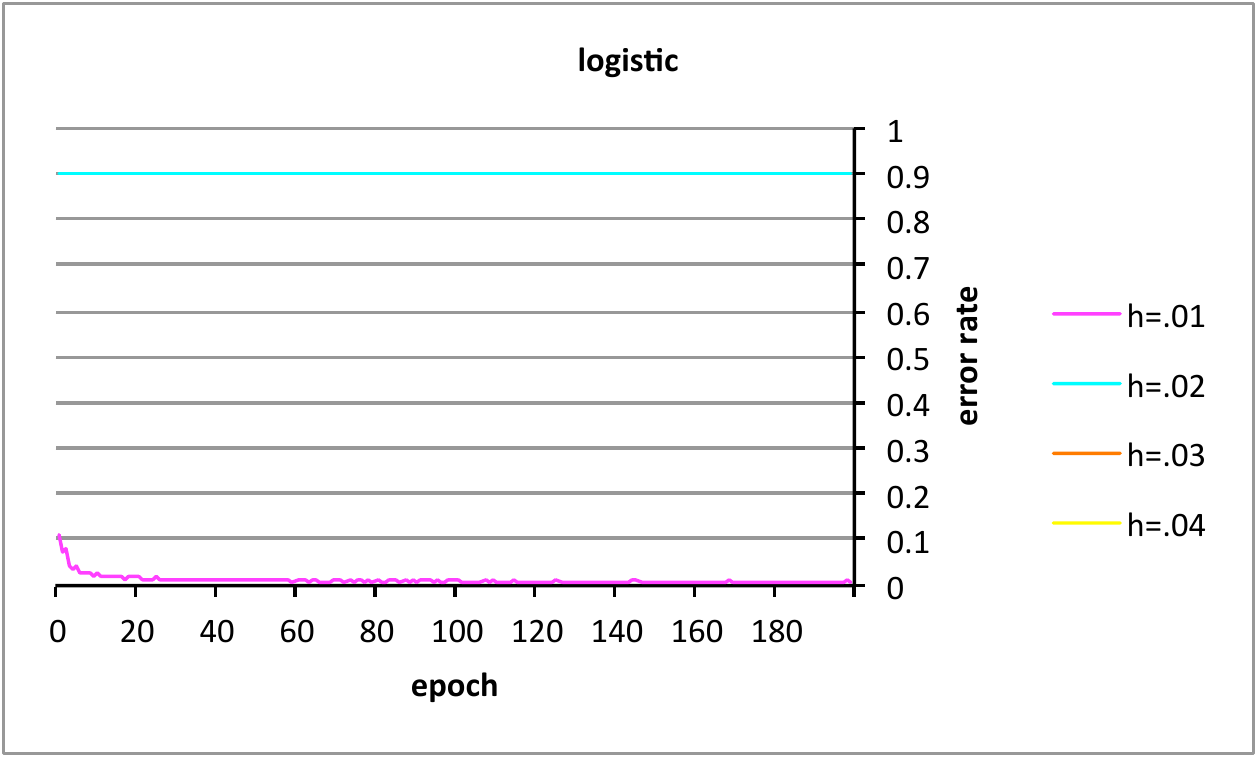}}
(c)\parbox{.93\linewidth}{\includegraphics[width=.92\linewidth]{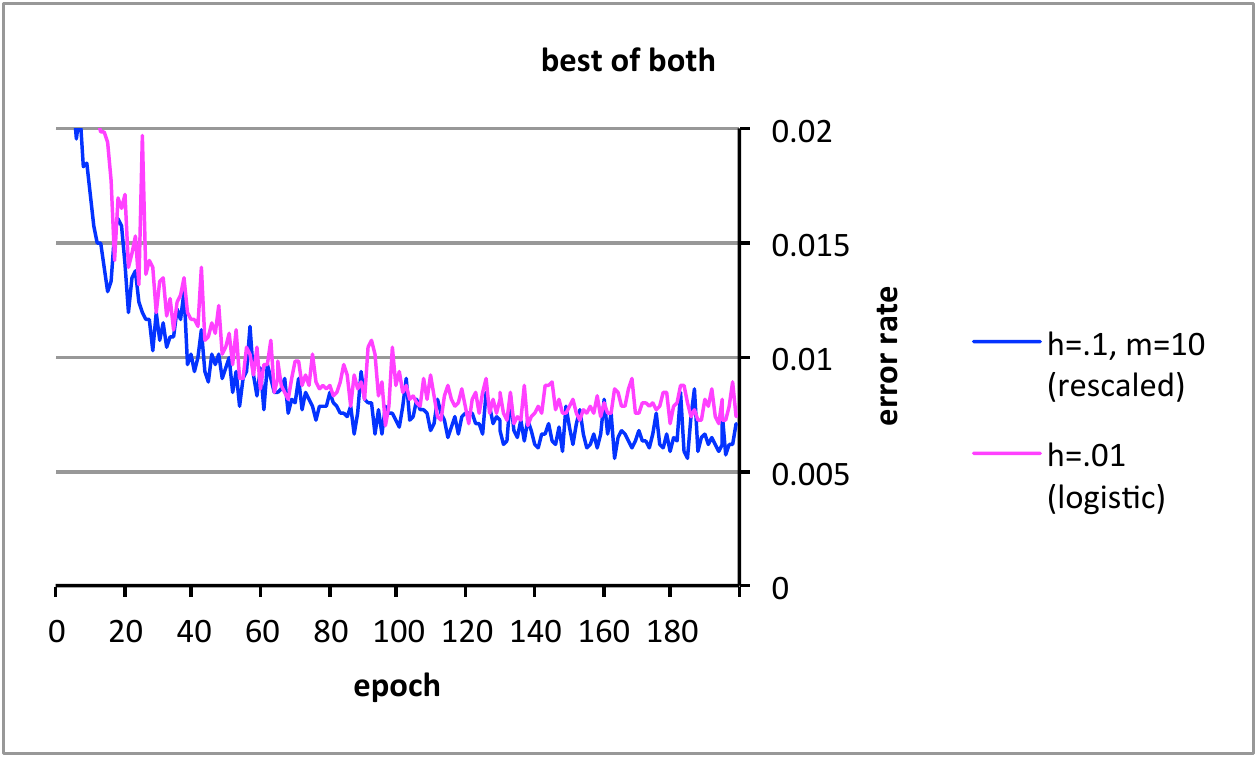}}
\end{flushright}
\caption{MNIST --- smaller architecture}
\label{mfigp}
\end{figure}

\begin{figure}
\begin{flushright}
(a)\parbox{.93\linewidth}{\includegraphics[width=.92\linewidth]{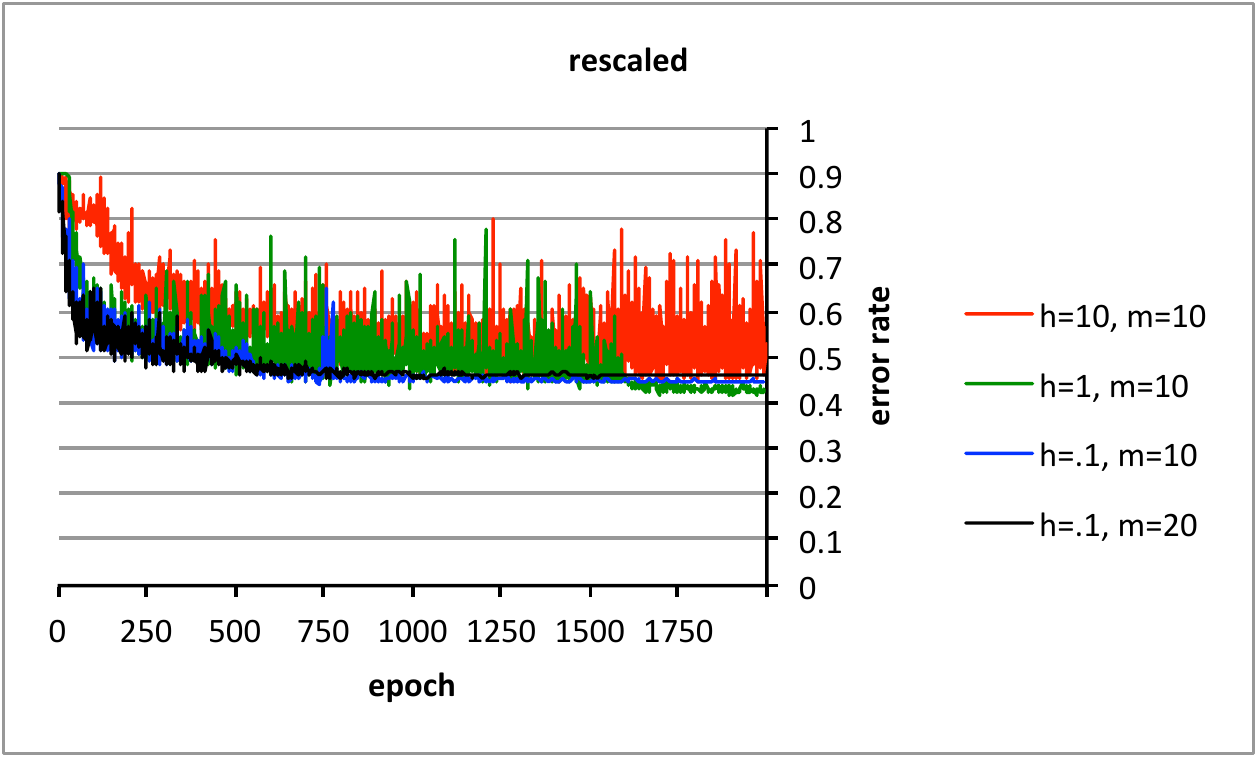}}
(b)\parbox{.93\linewidth}{\includegraphics[width=.92\linewidth]{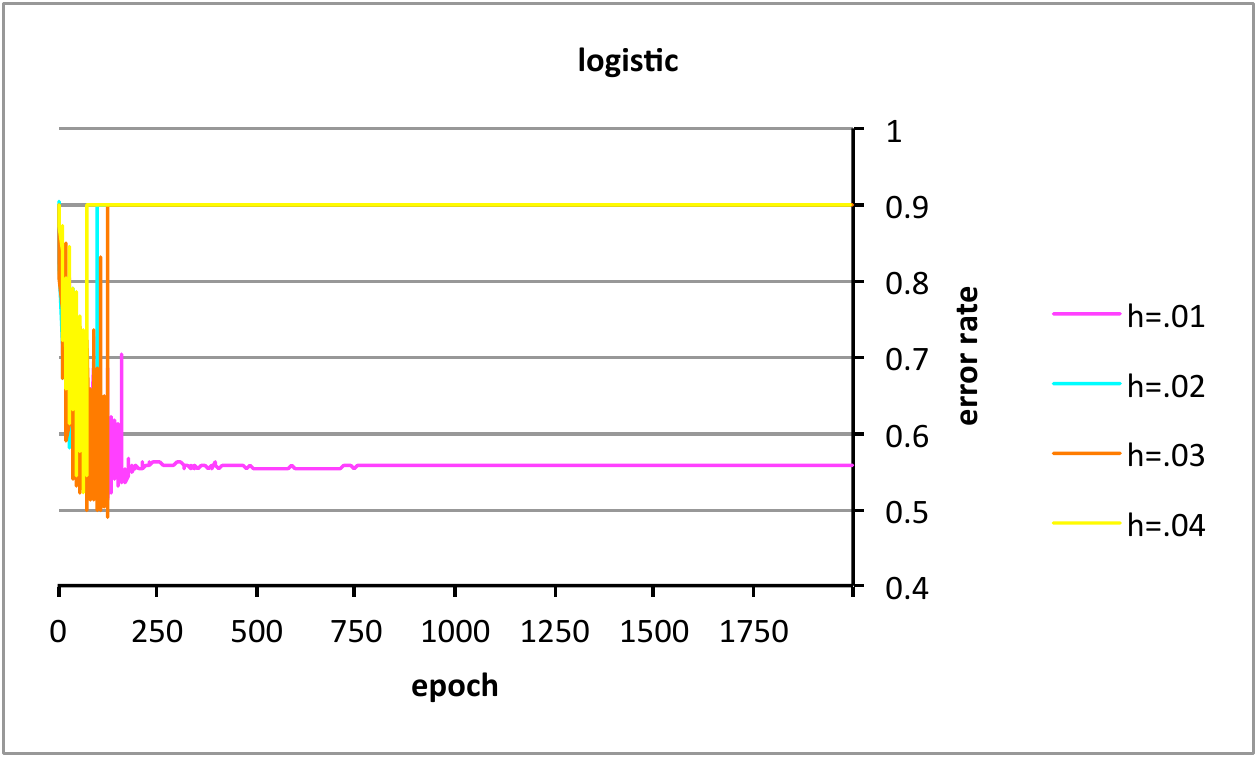}}
(c)\parbox{.93\linewidth}{\includegraphics[width=.92\linewidth]{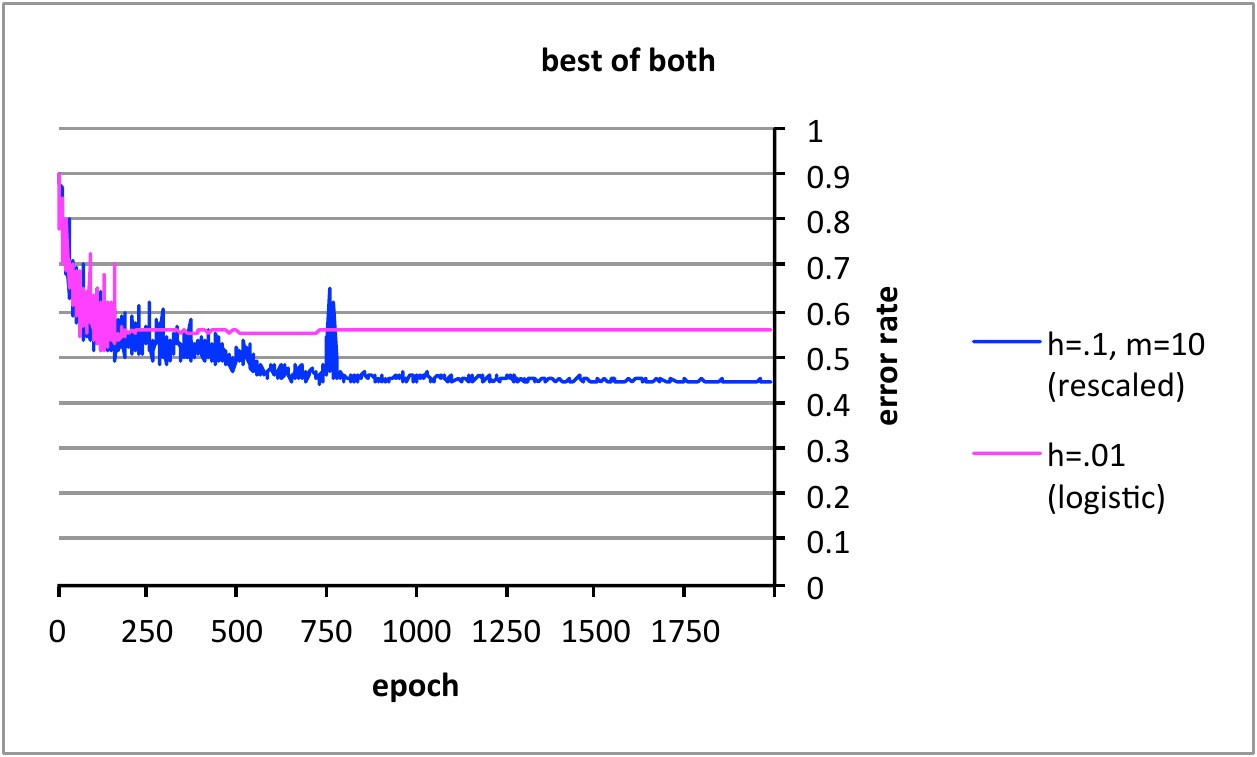}}
\end{flushright}
\caption{ImageNet subset --- smaller architecture}
\label{ifigp}
\end{figure}

Figures~\ref{cfig}--\ref{ifigp} plot the accuracies attained by
the different schemes for classification while varying $h$ from~(\ref{update})
($h$ is the ``learning rate,'' as well as the length of the learning step
relative to the magnitude of the gradient)
and varying the dimension $m$ of the space containing the simplex targets;
$m$ is the number of rows in $A$ from~(\ref{y}) and~(\ref{update}).
In each figure, the top panel --- that labeled ``(a)'' and ``rescaled'' ---
plots the error rates for classification using rescaling,
with the targets being the vertices on the hypersphere of a regular simplex;
the middle panel --- that labeled ``(b)'' and ``logistic'' ---
plots the error rates for classification using multinomial logistic regression;
the bottom panel --- that labeled ``(c)'' and ``best of both'' ---
plots the error rates for the best-performing instance from the top panel (a)
together with the best-performing instance from the middle panel (b). 
All error rates refer to performance on the test set.
The label ``epoch'' for the horizontal axes refers, as usual,
to the number of training sweeps through the data set,
as reviewed in the coming paragraph.

As recommended by~\citet{lecun-bottou-bengio-haffner},
we learn via (minibatched) stochastic gradient descent,
with 100 samples per minibatch; rather than updating the parameters
being learned for randomly selected individual images from the training set
exactly as in Section~\ref{new},
we instead randomly permute the training set and partition this permuted set
of images into subsets of 100, updating the parameters simultaneously
for all 100 images constituting each of the subsets (known as ``minibatches''),
processing the series of minibatches in series.
Each sweep through the entire training set is known as an ``epoch.''
The horizontal axes in the figures count the number of epochs.

In the experiments of the present section, the accuracies attained using
the scale-invariant classification stage are comparable to (if not better than)
those attained using the usual multinomial logistic regression.
Running the experiments with several different random seeds produces entirely
similar results.
The scale-invariant classification stage is stable for all values of $h$,
that is, for all learning rates.

\section{Conclusion}
\label{conclusion}
Combining [1] a convolutional network that is equivariant
to scalar multiplication,
[2] a classification stage that is invariant to scalar multiplication,
and [3] the rescaled learning of the last paragraph of Section~\ref{new}
fully realizes and leverages invariance to scalar multiplication.
This combination is more robust to outliers
(including obviously mislabeled data) than
the standard multinomial logistic regression ``softmax'' classification scheme,
results in marginally better errors on several standard test sets,
and fits/learns/trains precisely at the user-specified rate,
all while costing about the same computationally.
The attained invariance is clean and convenient --- a good goal all on its own.

\subsubsection*{Acknowledgments}

We would like to thank L\'eon Bottou and Rob Fergus
for critical contributions to this project.
The reviewers also helped immensely in improving the paper.

\begin{table}[h]
\caption{CIFAR-10 and MNIST, larger (\dag=1 for grayscale MNIST;
\dag=3 for full-color CIFAR-10)}
\label{standard}
\vspace{.5em}
\begin{tabular}{cccccc}
Stage & Input images & Output images & Kernel size
& Input image size & Output image size \\\hline
 first & \dag &   16 & $3 \times 3$ & $32 \times 32$ & $15 \times 15$ \\
second &   16 &  128 & $2 \times 2$ & $15 \times 15$ & $7 \times 7$ \\
 third &  128 & 1024 & $2 \times 2$ &   $7 \times 7$ & $3 \times 3$
\end{tabular}
\end{table}

\begin{table}[h]
\caption{CIFAR-10 and MNIST, smaller (\dag=1 for grayscale MNIST;
\dag=3 for full-color CIFAR-10)}
\label{standardp}
\vspace{.5em}
\begin{tabular}{cccccc}
Stage & Input images & Output images & Kernel size
& Input image size & Output image size \\\hline
 first & \dag &  16 & $3 \times 3$ & $32 \times 32$ & $15 \times 15$ \\
second &   16 &  64 & $2 \times 2$ & $15 \times 15$ & $7 \times 7$ \\
 third &   64 & 256 & $2 \times 2$ &   $7 \times 7$ & $3 \times 3$
\end{tabular}
\end{table}

\begin{table}[h]
\caption{ImageNet subset, larger (full color, with 3 color channels)}
\label{subset}
\vspace{.5em}
\begin{tabular}{cccccc}
Stage & Input images & Output images & Kernel size
& Input image size & Output image size \\\hline
 first &   3 &   16 & $5 \times 5$ & $128 \times 128$ & $62 \times 62$ \\
second &  16 &   64 & $3 \times 3$ &   $62 \times 62$ & $30 \times 30$ \\
 third &  64 &  256 & $3 \times 3$ &   $30 \times 30$ & $14 \times 14$ \\
fourth & 256 & 1024 & $3 \times 3$ &   $14 \times 14$ &   $6 \times 6$
\end{tabular}
\end{table}

\begin{table}[!h]
\caption{ImageNet subset, smaller (the smaller number is italicized
         in this table)}
\label{subsetp}
\vspace{.5em}
\begin{tabular}{cccccc}
Stage & Input images & Output images & Kernel size
& Input image size & Output image size \\\hline
 first &   3 &        16 & $5 \times 5$ & $128 \times 128$ & $62 \times 62$ \\
second &  16 &        64 & $3 \times 3$ &   $62 \times 62$ & $30 \times 30$ \\
 third &  64 &       256 & $3 \times 3$ &   $30 \times 30$ & $14 \times 14$ \\
fourth & 256 & {\it 256} & $3 \times 3$ &   $14 \times 14$ &   $6 \times 6$
\end{tabular}
\end{table}

\newpage

\bibliography{arxiv}
\bibliographystyle{iclr2016_conference}

\end{document}